\documentclass[letterpaper, 10 pt, conference]{IEEEtran}  

\IEEEoverridecommandlockouts                              

\usepackage{graphics} 
\usepackage{caption}
\usepackage{amsmath} 
\usepackage{amssymb}  
\usepackage{graphicx}
\usepackage{leftidx}
\usepackage{subfig}
\usepackage[dvipsnames]{xcolor}
\usepackage{algorithm}
\usepackage{algorithmic}
\usepackage{comment}
\usepackage{blindtext}
\usepackage[margin=1in]{geometry}
\usepackage{multirow}
\usepackage{multicol}
\usepackage{xcolor}
\usepackage{tabularray}
\usepackage{caption}
\usepackage{subcaption}
\usepackage{booktabs}
\usepackage[section]{placeins}
\usepackage{enumitem}

\def\R{\mathbb{R}}

\usepackage{xcolor}
\newcommand{\newedit}[1]{{#1}}

\allowdisplaybreaks

\title{\LARGE \bf Cascade IPG Observer for Underwater Robot State Estimation}

\author{Kaustubh Joshi, Tianchen Liu, Nikhil Chopra
\thanks{This work is supported by USDA NIFA Sustainable Agricultural Systems (SAS) Program (Award Number: 20206801231805).}%
\thanks{The authors are with the Department of Mechanical Engineering and Institute of Systems Research, University of Maryland, College Park, MD 20742, USA. {\tt \{kjoshi, tianchen, nchopra\}@umd.edu}.}
\thanks{Corresponding Author: \texttt{kjoshi@umd.edu}}}

\begin{document}

\maketitle
\thispagestyle{empty}
\pagestyle{empty}

\begin{abstract}

This paper presents a novel cascade nonlinear observer framework for inertial state estimation. It tackles the problem of intermediate state estimation when external localization is unavailable or in the event of a sensor outage. The proposed observer comprises two nonlinear observers based on a recently developed iteratively preconditioned gradient descent (IPG) algorithm. It takes the inputs via an IMU preintegration model where the first observer is a quaternion-based IPG. The output for the first observer is the input for the second observer, estimating the velocity and, consequently, the position. The proposed observer is validated on a public underwater dataset and a real-world experiment using our robot platform. The estimation is compared with an extended Kalman filter (EKF) and an invariant extended Kalman filter (InEKF). Results demonstrate that our method outperforms these methods regarding better positional accuracy and lower variance.

\end{abstract}

\section{Introduction}
\begin{figure}[t]
    \centering
    \includegraphics[width=\columnwidth]{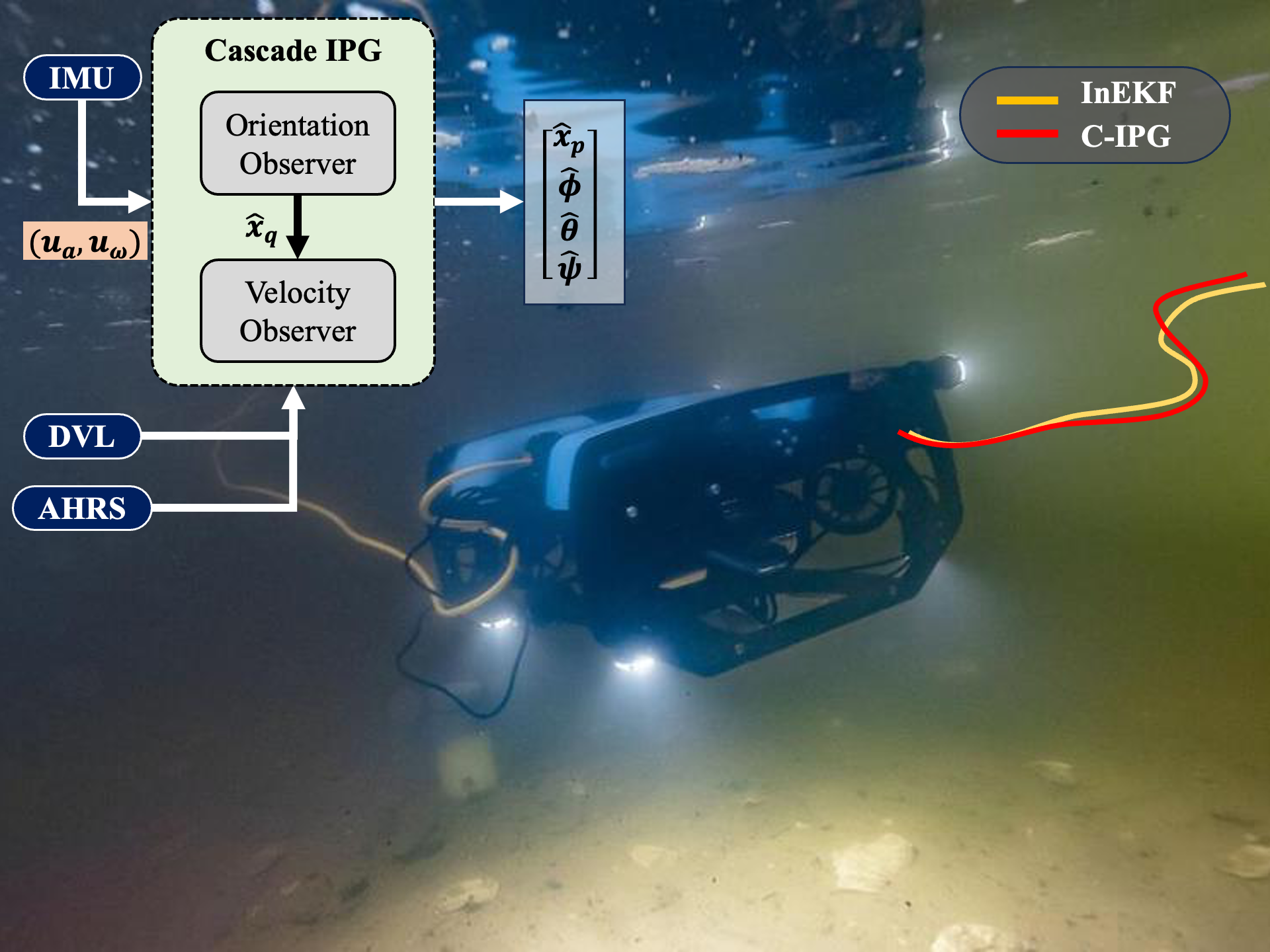}
    \caption{An underwater vehicle in operation and an overview of the cascade IPG (C-IPG) for pose estimation using various sensors on the ROV. An example of the estimated trajectories through implementing C-IPG and InEKF is shown in red and yellow (respectively). The image shows an experiment in the Chesapeake Bay (MD, USA), where turbid water affects visibility. Hence, accurate pose estimation by inertial sensors is critical.}
    \label{fig:beauty_shot}
\end{figure}

The interest in using underwater robots for surveying and exploring coastal environments for fishing and aquaculture has significantly increased in recent years \cite{li2018research}. Environmental and resource monitoring are crucial for enhancing efficiency and yields in the aquaculture industry. Utilizing underwater robots can make these tasks more innovative, cost-effective, and sustainable. Robot localization is pivotal in identifying areas conducive to high application yields.

However, water attenuates electromagnetic waves, rendering GPS and radar sensing underwater impractical. Current underwater robot localization and navigation methods predominantly rely on visual cues \cite{paull2013auv, lin2023uivnav}. Although vision-based algorithms have proven effective underwater \cite{zhang2022visual}, they tend to be computationally costly. Moreover, coastal environments, including aquaculture farms, frequently experience high turbidity, which adversely affects camera image quality \cite{davies2001turbidity, operations2003literature}. This diminishes the utility of visual data due to reduced visibility and poor illumination. While sonar and ultra-short baseline (USBL) can be alternatives \cite{bayat2015range, su2020review, gonzalez2020autonomous}, they come with high costs and technical challenges.

To address these challenges, our solution focuses on an inertial navigation framework using an Inertial Measurement Unit (IMU), Doppler Velocity Log (DVL), and Attitude Heading Reference Sensor (AHRS) to mitigate sensor outages. This paper presents a novel inertial state estimation framework that employs a cascade nonlinear observer, building upon our previous work on the IPG observer \cite{chak2023obsv}. We utilize the IMU preintegration method to tackle the dynamic challenges inherent in underwater vehicle systems. In summary, the key contributions of this paper are as follows:
\begin{itemize}[leftmargin=0.35cm]
\item We develop a novel cascade IPG (referred to as \textit{C-IPG}) observer for pose estimation of a 6-DOF system using IMU, DVL, and AHRS sensor readings. IMU preintegration eliminates the need for a dynamics model of the robot platform, enabling a model-free approach.
\item The performance of our methodology is validated using sensor data from (i) a real-world public dataset and (ii) our underwater robot platform. We implement the algorithm and compare its performance with filter-based methods. 
\end{itemize}

The paper is structured as follows: Section II concisely reviews related work. Section III outlines the necessary theoretical background. Section IV introduces the cascade IPG observer and its application. Section V validates the framework using a public dataset and our platform. Finally, Section VI concludes with final remarks and outlines future work.

\section{Related Work}
The localization problem for underwater vehicles is more challenging than other types of autonomous vehicles (e.g., ground or aerial vehicles) due to various uncertain environmental factors, and many solution approaches have been developed in recent decades. One main category is the simultaneous localization and mapping (SLAM)-based approaches. The applications of visual SLAM methods for underwater scenarios are addressed in~\cite{kim2013vslam_hull,carrasco2016stereo,weidner2017underwater, joshi2023sm, joshi2019experimental} when the camera data is usable for specific underwater environments. Additionally, acoustic sensor data can be incorporated into the pipeline for improved performance, as seen in~\cite{leutenegger2015, rahman2018sonar, vargas2021uw_vslam}, etc. However, all the above methods require vision or sonar cues. Gathering all data may not be possible due to factors like low visibility, limited power supply, computational power, etc. A second category is to employ nonlinear observers for estimating the states of a system (as described later in Eqs.~\eqref{eq:sys_dynamics},~\eqref{eq:sys_obs}). Localization can be regarded as a state estimation problem since the vehicle's pose is included in the states. For unmanned marine vehicles, including unmanned surface vehicles (USV) and unmanned underwater vehicles (UUV), nonlinear observers are designed for position estimation~\cite{fossen1999passive}, velocity estimation~\cite{liu2004nonlinear}, pose-inertial fusion \cite{lopez2023contracting}, etc. The cooperative localization problem is considered in~\cite{papadopoulos2010cooperative}. Recently, an InEKF-based localization approach for underwater vehicles using IMU, DVL, and pressure sensors has been developed in~\cite{potokar2021inekf_nav}. However, the approach was only tested using simulated sensor data, and the reader is referred to \cite{joshi20243d} for insights into implementing it for real-world experiments.

In this paper, we employ a recently developed novel approximate Newton observer, named IPG observer~\cite{chak2023obsv}, in a cascade framework for the state estimation problems of nonlinear systems and validate it on the localization of underwater vehicles. A combined use of IMU, DVL, and AHRS has been deployed. Additionally, state estimation models require knowledge of system dynamics, which can be challenging to obtain since underwater vehicles have dynamic uncertainty due to changing environmental conditions \cite{fossen1994guidance}. Hence, we incorporate IMU preintegration methods to get the resultant motion on the system directly. 
\newedit{In summary, we propose a model-free inertial observer (C-IPG) to bridge the research gap.}

\section{Theoretical Background}

\subsection{Notations: }

The notations used in this paper are listed as follows.

\textit{Algebra: }$\mathbb{R}$ denotes the set of real numbers. $\mathbb{R}^n$ is a set of n-dimensional real-valued vectors, $\mathbb{H}$ denotes Hamiltonian algebra, $\mathbf{I}$ denotes an identity matrix, $R$ denotes a rotation matrix, and $\otimes$ denotes the tensor product.

\textit{Reference Frames:} The proposed observer will consider three frames of reference: i) a global frame denoted as $(\cdot)^g$, centered at the Earth's center, ii) a local navigation frame $(\cdot)^n$ centered at a local fixed point (from which the robot starts moving) where the experiment is taking place, and iii) a body-fixed frame $(\cdot)^b$ onboard the robot in a NED (North-East-Down) configuration.

\subsection{Problem Description}
At time instant $k$, consider a discrete-time system as follows:
\begin{align}
	\mathbf{x}_{\{\textbf{s}\}, k+1} & = f(\mathbf{x}_{\{\textbf{s}\}, k}, \mathbf{u}_{\{\textbf{c}\},k}) + \varepsilon_{\mathbf{x}_{\{\textbf{s}\},k}}, \label{eq:sys_dynamics} \\
	\mathbf{z}_{\{\textbf{o}\},k} & = h(\mathbf{x}_{\{\textbf{s}\}, k}) + \varepsilon_{\mathbf{z}_{\{\textbf{o}\},k}}, \label{eq:sys_obs} \quad (k = 1, \dots, T),
\end{align}
where $f$ and $h$ denote the state propagation function and state-to-observation mapping function, respectively. $\mathbf{x}_k \in \mathbb{R}^n$ is the state $\{\textbf{s}\}$ ({\it position}, {\it velocity} or {\it quaternion}) with the corresponding dimension $n$, $\mathbf{u}_{\{\textbf{c}\},k} \in \mathbb{R}^m$ denotes the control input from the controller $\{\textbf{c}\}$ ({\it acceleration} or {\it angular velocity}) with the corresponding dimension $m$, and $\mathbf{z}_{\{\textbf{o}\},k} \in \mathbb{R}^p$ is the measurement from the sensor $\{\textbf{o}\}$ ({\it DVL} or {\it AHRS}) with the corresponding dimension $p$. $\varepsilon_{\mathbf{x}_{\{\textbf{s}\},k}} \in \mathbb{R}^n$ and $\varepsilon_{\mathbf{z}_{\{\textbf{o}\},k}} \in \mathbb{R}^p$ are Gaussian noise vectors for disturbances and measurement noises. 

Specifically for our underwater vehicle state estimation problem, the state variables include the position $\mathbf{x}_\mathbf{p}^n \in \mathbb{R}^3$, the linear velocity $\mathbf{x}_\mathbf{v}^n \in \mathbb{R}^3$, and the orientation, represented as a unit quaternion $\mathbf{x}_\mathbf{q}^n \in \mathbb{H}$. \newedit{While implementing the observer, we first represent the position in the local frame. After processing, the position is transformed to the global frame for the final result.} For brevity, \newedit{the states in local frame} are hereon denoted as $\mathbf{x}_\mathbf{p}$, $\mathbf{x}_\mathbf{v}$ and $\mathbf{x}_\mathbf{q}$, respectively. For the control input variables, $\mathbf{u}^b_\mathbf{a} \in \mathbb{R}^3$ is the linear acceleration and $\mathbf{u}^b_\mathbf{\omega} \in \mathbb{R}^3$ is the angular velocity from the IMU readings in the body frame, referred to as $\mathbf{u}_\mathbf{a}$ and $\mathbf{u}_\mathbf{\omega}$ in this paper. Finally, $\mathbf{z}^n_{D} \in \mathbb{R}^3$ denotes the linear velocity measurements from DVL, and $\mathbf{z}^n_{A} \in \mathbb{H}$ denotes the quaternion measurements from AHRS. We use $\mathbf{z}_D$ and $\mathbf{z}_A$ to represent them in the remainder of the paper.

\subsection{IPG Observer}
In this paper, the IPG observer~\cite{chak2023obsv} estimates the states in the IMU preintegration method. In this section, we briefly introduce the approach. A more detailed description and the convergence proof can be found in~\cite{chak2023obsv} and~\cite{chak2024convergence}, respectively.

Consider the discrete-time nonlinear system defined as in Eqs.~\eqref{eq:sys_dynamics} and \eqref{eq:sys_obs}. At the $k$-th time instant ($k \geq N$), we define $f^{\mathbf{u}_{\{\textbf{c}\}, k}} := f(\mathbf{x}_{\{\textbf{s}\}, k}, \mathbf{u}_{\{\textbf{c}\},k})$. Let $\mathbf{\mathrm{Z}}_{\{\textbf{o}\},k} \in \mathbb{R}^{Np}$ and $\mathbf{\mathrm{U}}_{\{\textbf{c}\},k} \in \mathbb{R}^{(N-1)m}$ denote the concatenating column vectors of the past $N$ consecutive measurements from $\{\textbf{o}\}$ and $(N-1)$ inputs from $\{\textbf{c}\}$, respectively, i.e.,
{\small
\begin{align}
	&\mathbf{\mathrm{U}}_{\{\textbf{c}\},k} =  \left[\mathbf{u}_{\{\textbf{c}\}, k-N+1}^T, \dots , \mathbf{u}_{\{\textbf{c}\}, k-1}^T \right]^T, \\
	&\mathbf{\mathrm{Z}}_{\{\textbf{o}\}, k} =  \left[\mathbf{z}_{\{\textbf{o}\}, k-N+1}^T, \dots, \mathbf{z}_{\{\textbf{o}\}, k}^T \right]^T \nonumber \\
	&= \begin{bmatrix}
		h(\mathbf{x}_{\{\textbf{s}\}, k-N+1}) \\
		h \circ f^{\mathbf{\mathbf{u}}_{\{\textbf{c}\}, k-N+1}}(\mathbf{x}_{\{\textbf{s}\}, k-N+1}) \\
		\vdots \\
		h \circ f^{\mathbf{u}_{\{\textbf{c}\}, k-1}} \circ \dots \circ f^{\mathbf{u}_{\{\textbf{c}\}, k-N+1}}(\mathbf{x}_{\{\textbf{s}\}, k-N+1})
	\end{bmatrix} \nonumber \\
 & +
	\begin{bmatrix}
		\varepsilon_{\mathbf{z}_{\{\textbf{o}\},k-N+1}} \\
		\varepsilon_{\mathbf{z}_{\{\textbf{o}\},k-N+2}} \\
		\vdots \\
		\varepsilon_{\mathbf{z}_{\{\textbf{o}\}, k}}
	\end{bmatrix} \quad :=  H^{\mathbf{\mathrm{U}}_{\{\textbf{c}\},k}}(\mathbf{x}_{\{\textbf{s}\}, k-N+1}) + \varepsilon_{\{\textbf{o}\},k}, \label{eqn:meas_vec}
\end{align}
}
where `$\circ$' denotes composition of functions. For each $k \geq N$, an estimate of $\mathbf{x}_{\{\textbf{s}\},k}$ is obtained by solving for $\mathbf{x}_{\{\textbf{s}\}, k-N+1}$ in~\eqref{eqn:meas_vec} and propagating the obtained solution of $\mathbf{x}_{\{\textbf{s}\},k-N+1}$ forward by $N$ time instants using the dynamics function $f$. 

The IPG observer can be regarded as a Newton-type observer. The original Newton observer~\cite{moraal1995observer} requires computing the inverse (or pseudo-inverse) of the Jacobian matrix of $H^{\mathbf{U}_{\{\textbf{c}\},k}}$. In the IPG observer~\cite{chak2023obsv}, instead of calculating the inverse of the Jacobian, its approximation is obtained by an iterative process in the form of a preconditioner matrix. The iterations are implemented with updates of the current state estimates over a sliding window (moving horizon). Specifically, at each instant $k \geq N$, the IPG observer applies $d$ iterations indexed by $i=0,\ldots,d-1$ in an inner loop. In the $i$-th inner loop, an estimate $\zeta_{\{\textbf{s}\},k}^i \in \R^n$ of $\mathbf{x}_{\{\textbf{s}\},k-N+1}$ and a preconditioner matrix $K_k^{i+1} \in \R^{n \times n}$ are maintained. Before the inner loops start, the iterates $\zeta_{\{\textbf{s}\},N}^0 \in \R^n$ and $K_N^0 \in \R^{n \times n}$ are initialized, and the number of inner loop iterations $d$ is selected.

For the $k$-th time instant, the IPG observer approach is processed as follows, 

\noindent \textbf{Step I.} At time instant $k$, $d$ iterations of inner loops are executed for updating the estimate $\zeta_{\{\textbf{s}\}, k}^0$ and preconditioner $K_k^0$. For each inner iteration $i=0,\ldots,d-1$, let 
$J_k^i$ denote the Jacobian of $H^{\mathbf{\mathrm{U}}_{\{\textbf{c}\}, k}}$ evaluated at $\zeta_{\{\textbf{s}\}, k}^i$, then $K_k^i$ and $\zeta_{\{\textbf{s}\}, k}^i$ are updated as,
{\small
\begin{align}
    & K^{i+1}_k =  K^{i}_k - \alpha^i \left(\left(J_k^i\right)^T J_k^i K^{i}_k - \mathbf{I} \right), \label{eqn:K_rec} 
    \\
    & \zeta^{i+1}_{\{\textbf{s}\}, k} = \zeta^{i}_{\{\textbf{s}\}, k} - \delta^i  K^{i}_k \left(J_k^i\right)^T \left( H^{\mathbf{\mathrm{U}}_{\{\textbf{c}\}, k}}(\zeta^i_{\{\textbf{s}\}, k}) - \mathbf{\mathrm{Z}}_{\{\textbf{o}\}, k} \right), \label{eqn:ipg_rec}
\end{align}
}
for $i=0,\ldots,d-1$, where the parameters $\alpha^i$ and $\delta^i$ are step size values for each inner loop. 

\noindent \textbf{Step II.} The value of $\zeta_{\{\textbf{s}\}, k}^d$, obtained after $d$ inner iterations, is used to compute the estimate $\hat{x}_{\{\textbf{s}\}, k}$ as
{\small
\begin{align}
    \hat{\mathbf{x}}_{\{\textbf{s}\}, k} & = f^{\mathbf{u}_{\{\textbf{c}\}, k-1}} \circ \ldots \circ f^{\mathbf{u}_{\{\textbf{c}\}, k-N+1}}(\zeta_{\{\textbf{s}\}, k}^d). \label{eqn:newton_est}
\end{align}
}

\noindent \textbf{Step III.} The initial estimate and preconditioner for the next time instant $k+1$ are obtained by propagating forward the values of $\zeta_{\{\textbf{s}\}, k}^d$ and $K_k^d$ from \textbf{Step I} as
{\small
\begin{align}
    \zeta_{\{\textbf{s}\}, k+1}^0 & = f^{\mathbf{u}_{\{\textbf{c}\}, k-N+1}}(\zeta_{\{\textbf{s}\}, k}^d), \quad K_{k+1}^0 = K_k^d, \label{eqn:ipg_init}
\end{align}
}
and the process \textbf{Steps I-III} repeats for time instant $k+1$.

\subsection{IMU Preintegration}

For a practical underwater vehicle platform, obtaining an accurate dynamical model can be challenging due to uncertain environmental conditions and the structural variations of the vehicle~\cite{fossen1994guidance}. Hence, an IMU-based state estimation, named IMU preintegration~\cite{forster2015imu}, is considered in this case. The IMU sensor provides high-frequency measurements for linear acceleration $\mathbf{u}_\mathbf{a}$ and the angular velocity $\mathbf{u}_\mathbf{\omega}$ in the body frame, along with their respective biases $\mathbf{b}_a , \mathbf{b}_{\omega} \in \mathbb{R}^3$. These measurements enable an IMU propagation model for the state estimation. In IMU preintegration, the acceleration and angular velocities are obtained from the IMU in the body frame and then corrected with the initial velocity and gravity. The method is formulated as follows \cite{sola2017quaternion},
{\small
\begin{align}
    \dot{\mathbf{x}}_\mathbf{p} &= \mathbf{x}_\mathbf{v}, \quad \dot{\mathbf{x}}_\mathbf{v} = R(\mathbf{x}_\mathbf{q})(\mathbf{u}_\mathbf{a} - \mathbf{b}_{a}) +  \mathbf{g}, \\
    \dot{\mathbf{x}}_\mathbf{q} &= \dfrac{1}{2} \mathbf{x}_\mathbf{q} \otimes \begin{bmatrix}
        0 \\
        \mathbf{u}_\omega - \mathbf{b}_{\omega}
    \end{bmatrix}, \label{eq: quat_int}
\end{align}
}
where $R(\mathbf{x}_\mathbf{q})$ represents the rotation matrix derived from the quaternion and $\mathbf{g}$ represents the gravity vector in the local frame. 
These equations are discretized with small time intervals when using the IPG observer.


\vspace{6pt}
In the next section, we introduce the proposed cascade IPG observer as a novel approach for the state estimation of underwater vehicles.

\begin{figure}[!tpb]
\includegraphics[width=\linewidth]{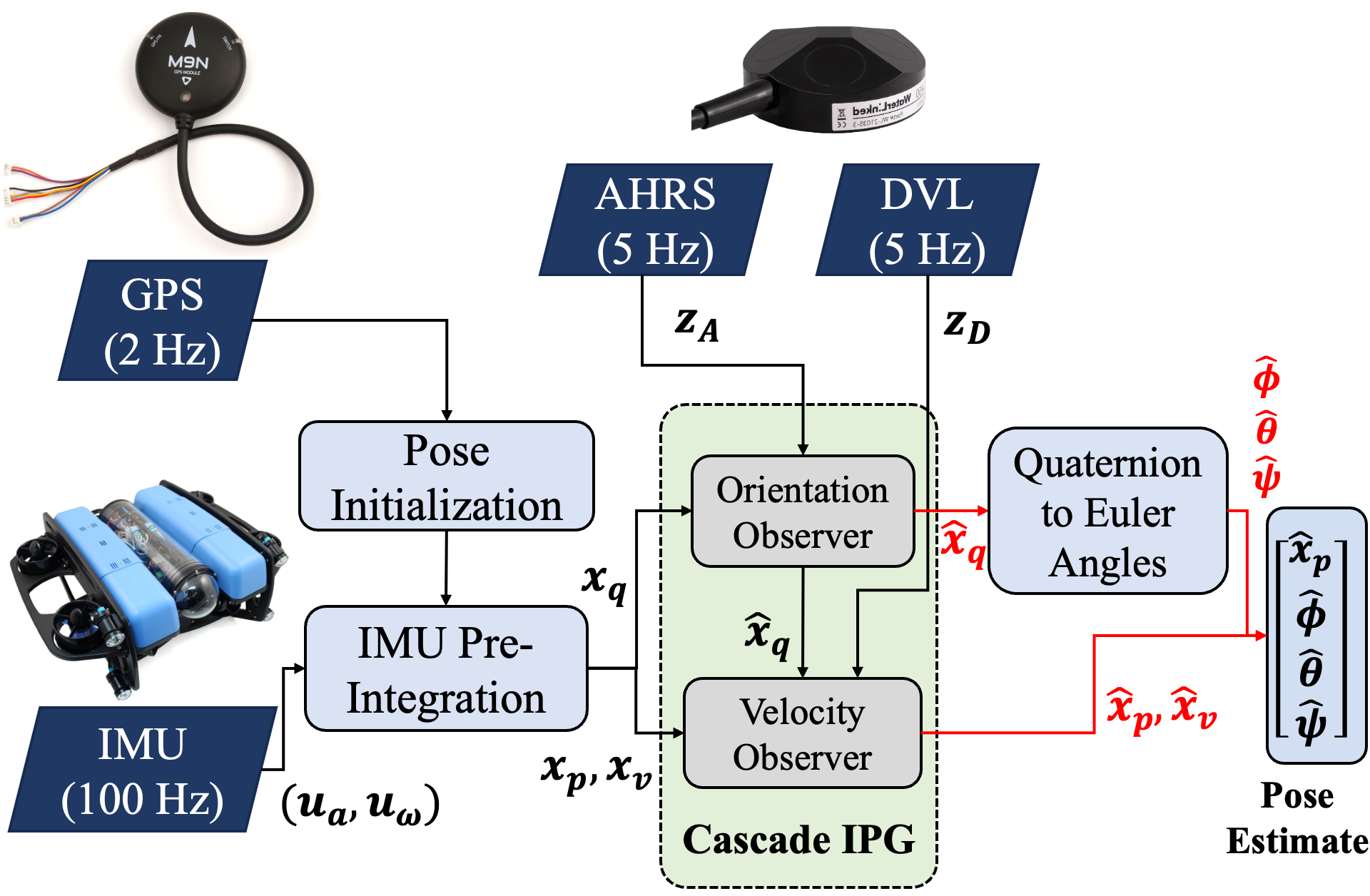}
\centering
\caption{Framework of the proposed cascade IPG observer.}
\label{fig:overview}
\end{figure}

\section{Proposed Approach}
The framework of the proposed cascade IPG observer approach is presented in Fig. \ref{fig:overview}. The framework involves a novel cascade structure, where we estimate the orientation first, and the predicted orientation estimate is used for calculating the velocity and, consequently, the robot's position. 
Using the IMU preintegration approach, we leverage measurements from onboard DVL and AHRS sensors to correct the drift. 

\begin{algorithm}[!t]
\caption{Cascade IPG Algorithm}\label{alg:1}
\begin{algorithmic}[1]
{\small
	\STATE Initialize estimates $\hat{\mathbf{x}}_\mathbf{p}$, $\hat{\mathbf{x}}_\mathbf{v}$, $\hat{\mathbf{x}}_\mathbf{q}$.
	\FOR{time instant $k \geq N$}
    \color{Cerulean}
    \STATE \textbf{Orientation Observer}
	\STATE Obtain vectors $U_{\omega, k}$ from IMU and $Z_{A,k}$ from AHRS
    {\small
        \STATE $\zeta^{0}_{\mathbf{q}, k} = f_\mathbf{q}^{\mathbf{u}_{\omega, k-N+1}}(\hat{\mathbf{x}}_\mathbf{q}, \mathbf{u}_\omega, \mathbf{b}_\omega)_{k-N+1}$ - \textit{see Eq. (\ref{eqn:quat_observer})}
    }
	\FOR{$i$ in $0:d-1$} 
	\STATE $K^{i+1}_k = K^{i}_k - \alpha^i \left( (J_k^i)^T J_k^i K^{i}_k - \mathbf{I} \right)$
        \STATE $\zeta^{i+1}_{\mathbf{q}, k} = \zeta^{i}_{\mathbf{q}, k} - \delta^i  K^{i}_k (J_k^i)^T  ( H^{\mathbf{\mathrm{U}}_{\omega, k}} - \mathbf{\mathrm{Z}}_{A,k} ) $
    \ENDFOR
    \STATE $\mathbf{\hat{x}}_\mathbf{q} =\zeta^{d}_{\mathbf{q}, k}$
    \color{Rhodamine}
    \STATE \textbf{Velocity Observer}
	\STATE Obtain vectors $U_{\mathbf{a}, k}$ from IMU and $Z_{D,k}$ from DVL
    {\small
        \STATE $\zeta^{0}_{\mathbf{v}, k} = f_\mathbf{v}^{\mathbf{u}_{\mathbf{a}, k-N+1}}(\hat{\mathbf{x}}_\mathbf{v}, \mathbf{u}_\mathbf{a}, \mathbf{b}_\mathbf{a}, \hat{\mathbf{x}}_\mathbf{q})_{k-N+1}$ - \textit{see Eq. (\ref{eqn:pos_from_vel})}
	}
    \FOR{$i$ in $0:d-1$} 
	\STATE $K^{i+1}_k = K^{i}_k - \alpha^i \left( (J_k^i)^T J_k^i K^{i}_k - \mathbf{I} \right)$
        \STATE $\zeta^{i+1}_{\mathbf{v}, k} = \zeta^{i}_{\mathbf{v}, k} - \delta^i  K^{i}_k (J_k^i)^T  ( H^{\mathbf{\mathrm{U}}_{\mathbf{a}, k}} - \mathbf{\mathrm{Z}}_{D,k} ) $
    \ENDFOR
    \STATE $\mathbf{\hat{x}}_\mathbf{v} = \zeta^{d}_{\mathbf{v}, k}$
    {\small
    \STATE $\mathbf{\hat{x}}_\mathbf{p} = \hat{\mathbf{x}}_\mathbf{p} + \mathbf{\hat{x}}_\mathbf{v}  \Delta t_k$
    }
    \ENDFOR
    }
\end{algorithmic}
\end{algorithm}

The detailed implementation of the proposed approach is summarized in Algorithm~\ref{alg:1}. Specifically, the approach consists of two major steps. The \textbf{first step} is to design the orientation observer. The underlying theory is to estimate the quaternion at every time instant. Quaternion integration establishes the relationship between the angular velocity and the resulting quaternion \cite{andrle2013geometric}. The angular velocity measured by the IMU is assumed constant between two consecutive time instants. \newedit{By discretizing Eq.~\eqref{eq: quat_int}}, the state space equation for the \textit{quaternion observer} is given in Eq.~\eqref{eqn:quat_observer}, where 
$\hat{\mathbf{x}}_\mathbf{q}$ is the estimated quaternion vector from the previous time instant.
{\small
\begin{align}
    \zeta^{0}_{\mathbf{q}, k} 
    & =  \Delta t_k \bigg( \dfrac{1}{2} \hat{\mathbf{x}}_{\mathbf{q},k-N+1} \otimes \begin{bmatrix}
        0 \\
        \mathbf{u}_\omega - \mathbf{b}_{\omega}
    \end{bmatrix} \bigg) \nonumber \\
    & := f_{\mathbf{q}}^{\mathbf{u}_{\omega, k-N+1}}(\hat{\mathbf{x}}_\mathbf{q}, \mathbf{u}_\omega, \mathbf{b}_\omega)_{k-N+1}.
    \label{eqn:quat_observer}
\end{align}
}
Here, the subscript $(k-N+1)$ means the variable values at the $(k-N+1)$-th time instant. The IPG observer is then employed to obtain an estimate for the quaternion as $\hat{\mathbf{x}}_\mathbf{q}$, which will be used in the next step.

The \textbf{second step} is to design the velocity (translation) observer, which depends on the rotation matrix $R$ from the estimated quaternion $\hat{\mathbf{x}}_\mathbf{q}$.
Together with the previously estimated velocity of the robot $\hat{\mathbf{x}}_\mathbf{v}$, the initial velocity estimate for the IPG observer of translation is calculated as 
{\small
\begin{align} \nonumber
     \zeta^{0}_{\mathbf{v}, k} 
     & = \hat{\mathbf{x}}_\mathbf{v} + \Delta t_k \big[ R(\hat{\mathbf{x}}_{\mathbf{q},k-N+1})(\mathbf{u}_{\mathbf{a},k-N+1} - \mathbf{b}_{a}) +  \mathbf{g} \big]\\ 
     & := f_\mathbf{v}^{\mathbf{u}_{\mathbf{a}, k-N+1}}(\hat{\mathbf{x}}_\mathbf{v}, \mathbf{u}_\mathbf{a}, \mathbf{b}_\mathbf{a}, \hat{\mathbf{x}}_\mathbf{q})_{k-N+1}, \label{eqn:pos_from_vel} 
\end{align}
}
After the IPG translation observer is computed, we update $\hat{\mathbf{x}}_\mathbf{v}$ and $\hat{\mathbf{x}}_\mathbf{p}$ by,
{\small
\begin{align}
    & \mathbf{\hat{x}}_\mathbf{v} = \zeta^{d}_{\mathbf{v}, k}, \quad \mathbf{\hat{x}}_\mathbf{p} = \hat{\mathbf{x}}_\mathbf{p} + \mathbf{\hat{x}}_\mathbf{v}  \Delta t_k.
\end{align}
}

\newedit{\textit{Remark:} Although the IPG observer algorithm is employed without much modification from~\cite{chak2023obsv}, the cascade structure enables a better estimate of orientation before the velocity observer, leading to improved state estimation results.}
Note that $\mathbf{b}_{\mathbf{a}}$ is generally obtained from an IMU, either via its built-in bias estimation method or through calibration, and $\mathbf{b}_{\omega}$ is estimated using a VQF \cite{laidig2023vqf} method. Including biases in the pipeline improves accuracy, but C-IPG can still work regardless of their inclusion. \newedit{Currently, the C-IPG observer has not been proven to be asymptotically stable, as the system is partially observable. Our future work will investigate the cascade structure of a fully observable system with stability analysis.} 








\section{Experimental Results}
The experimental validation is carried out on a publicly available dataset and our robotic system. We aim to validate the performance of C-IPG on these datasets and compare them with EKF and InEKF. 

\begin{figure}
    \centering
    \includegraphics[width=\columnwidth, trim = {0 0 0 4cm}, clip]{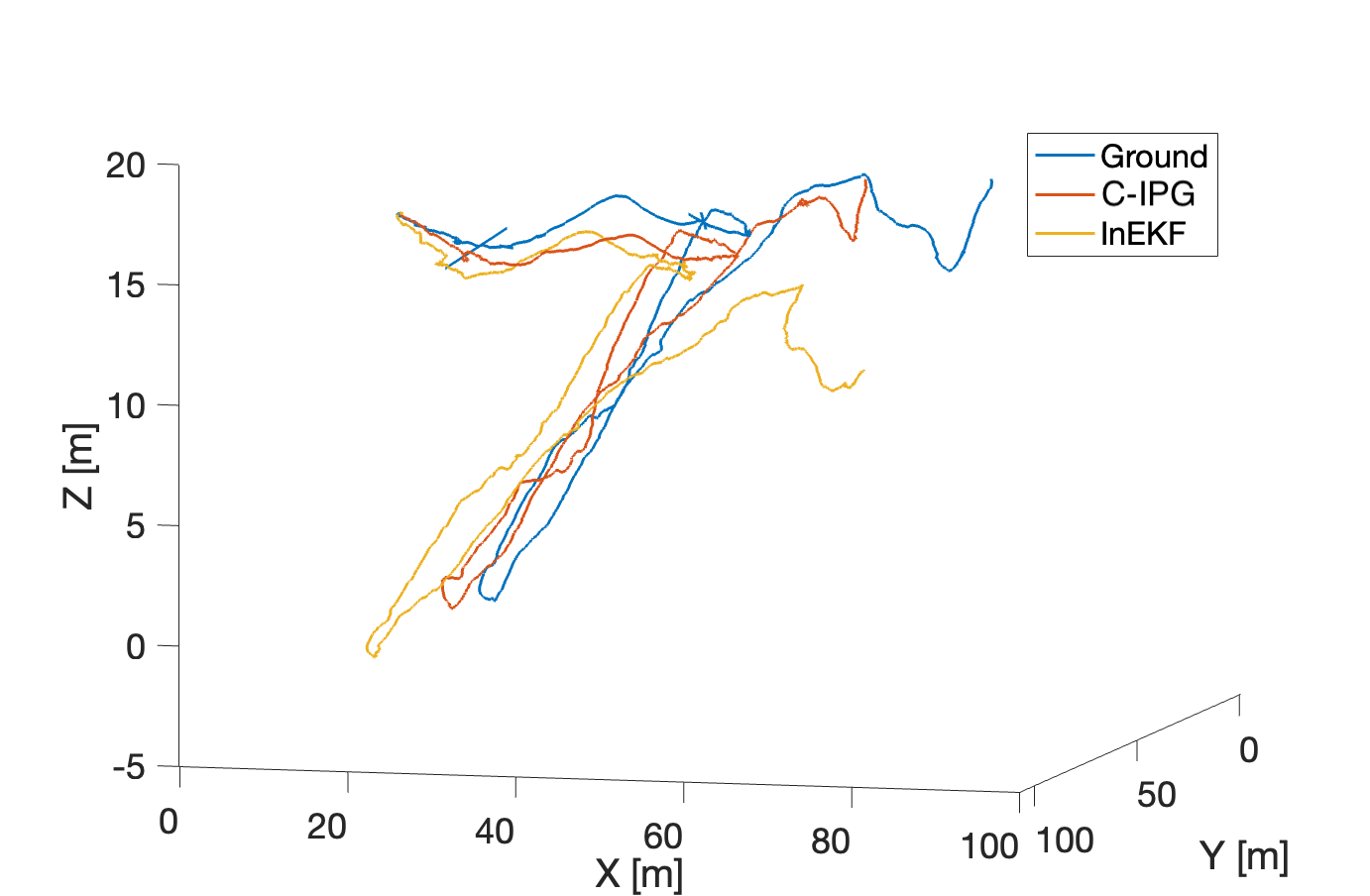}
    \caption{Comparison of 3D trajectory estimations of the vehicle from Girona dataset.}
    \label{fig:girona_traj}
\end{figure}

\begin{figure*}
    \centering
    \includegraphics[width=\textwidth, trim = {0 0 0 1cm}, clip]{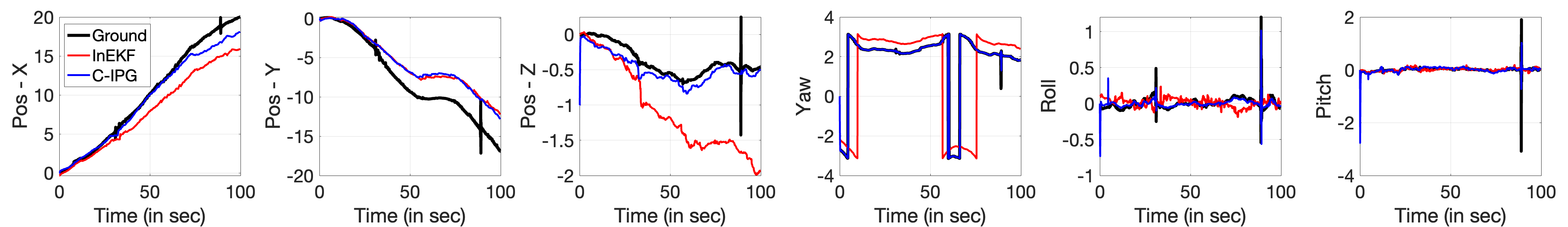}
    \caption{Comparison for individual states of the Girona dataset for first 100 seconds. The black line indicates the ground truth (contains occasional outliers resulting in spikes). The red and blue lines represent the states estimated by InEKF and C-IPG, respectively.}
    \label{fig:uuv_positions_girona}
\end{figure*}

\begin{figure*}
    \centering
    \includegraphics[width=\textwidth, trim = {0 0 0 1cm}, clip]{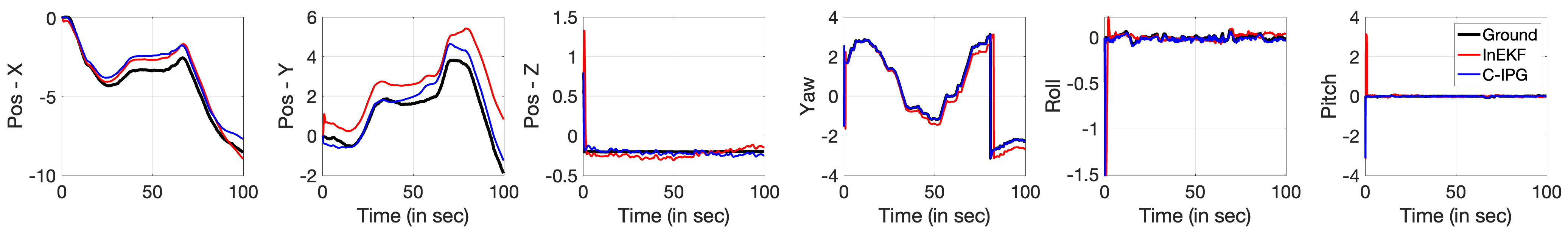}
    \caption{Comparison for individual states of the BlueROV2 for first 100 seconds. The black line indicates the ground truth. The red and blue lines represent the states estimated by InEKF and C-IPG, respectively.}
    \label{fig:uuv_positions_bluerov}
\end{figure*}

\begin{table*}
    \centering \scriptsize
    \caption{Comparison of Mean Absolute Error (MAE) for individual states, the overall Root Mean Square Error (RMSE), and the total runtime for EKF, InEKF, and C-IPG over various periods. The Total Error is computed as the RMSE of the MAEs obtained from all states, while the Total Variance is defined as the variance computed across the error values of all states.}
    \begin{tabular}{lcccccccccccl}\toprule
MAE for State & \multicolumn{3}{c}{Girona (100 s)} & \multicolumn{3}{c}{BlueROV2 (100 s)} & \multicolumn{3}{c}{Girona (1000 s)} & \multicolumn{3}{c}{BlueROV2 (300 s)}
\\\cmidrule(lr){2-4}\cmidrule(lr){5-7}\cmidrule(lr){8-10}\cmidrule(lr){11-13}
          & EKF  & InEKF & C-IPG    & EKF  & InEKF & C-IPG  & EKF  & InEKF & C-IPG & EKF  & InEKF & C-IPG\\\midrule
Pos - $\mathbf{x}$    & 1.4938 & 2.1038 & 0.7934 & 0.7698 & 0.4812 & 0.6101 & 5.4750 & 5.4396 & 5.3719 & 1.5410 & 1.2766 & 1.5887 \\
Pos - $\mathbf{y}$  & 2.5082 & 2.0726 & 2.2969 & 0.8997 & 1.2794 & 0.4175 & 7.9169 & 11.0765 & 6.1148 & 1.4022 & 2.6352 & 1.0375 \\
Pos - $\mathbf{z}$  & 0.3316 & 0.6836 & 0.0575 & 0.1967 & 0.0631 & 0.0286 & 2.2238 & 3.1354 & 1.2776 & 0.3060 & 0.0758 & 0.0947\\ \midrule
Total Error  & 10.4931 & 3.0828 & \textbf{2.3767} & 5.7409 & 1.5065 & \textbf{0.9012} & 38.2903 & 13.1961 & \textbf{8.2622} & 8.0703 & 3.0448 & \textbf{2.1286}\\
Total Variance  & 32.6206 & 3.8604 & \textbf{1.9686} & 7.6428 & 0.6042 & \textbf{0.0889} & 254.0047 & 25.5553 & \textbf{10.9253} & 26.0001 & \textbf{2.0545} & 2.1464\\\midrule
Runtime (in s)  & \textbf{2.7594} & 9.1428 & 27.9138 & \textbf{3.0306} & 11.0643 & 30.1871 & \textbf{26.9367} & 40.2762 & 253.4947 & \textbf{9.0348} & 19.5921 & 66.4363 \\\bottomrule
\end{tabular}
\label{tab:results}
\end{table*}

\subsection{Experiment with Underwater Cave Dataset}

The Girona underwater cave dataset \cite{girona2017} is a public dataset collected in an underwater environment. It comprises data from IMU, DVL, and AHRS, as well as visual and sonar data. The measurements from the DVL are crucial for validating an inertial navigation policy. We constructed a system model using IMU's control inputs, the linear velocities from the DVL, and orientations from the AHRS readings as measurements. We referenced our results against the odometry readings provided in the dataset. This analysis allows us to evaluate the algorithm's performance under conditions where vision (or other external sensor data) is corrupted or unavailable for localization. More details and descriptions of the sensors used in this dataset can be found in \cite{girona2017}. We evaluate and compare C-IPG with EKF and InEKF on the dataset's first 100 and 1000 seconds. \newedit{Note that all approaches can run for a longer period, but the focus of the tests is to show the validity of the proposed observer.} 

\subsection{Experiments with BlueROV2}

Our real-world experiments are carried out in the Chesapeake Bay at multiple sites as leases for oysters. The objective is to monitor water quality variability using sensors on the BlueROV2 platform~\cite{BlueROV2}, where the vehicle's localization is critical. One challenge is limited visibility during the experiments due to plumes generated by dredging vessels. The reader is referred to the supplemental video to interpret the experiments visually. 

The BlueROV2 Heavy Configuration is used to conduct the experiments. It has 8 T200 thrusters for motion in all 6 degrees of freedom (DOF). The onboard sensors include TDK InvenSense ICM-20602 (IMU) at 100 Hz, Waterlinked DVL-A50 (DVL) with an AHRS at 5 Hz, and a u-blox M9N GPS module at 2 Hz. 
GPS sensor is retrofitted 0.5 m above the ROV frame, and the measurements are used as the ground truth. The vehicle has a significantly lower cost than the one used in the Girona dataset, enabling the validation of C-IPG on low-cost systems. 

The initial heading of the ROV is observed from the onboard compass \newedit{and aligned~\cite{8593941} with the first few GPS readings}. The subsequent operation of the ROV results in fluctuations due to varying electromagnetic fields (EMF) in the electronics enclosure. The ROV was operated slightly below the surface at a depth of around 0.2m so that the ROV was completely submerged. Still, the GPS was always over the surface, enabling continuous transmission. The ROV was operated at low speeds ($<1$ m/s) in various paths in the bay, and all sensor data was collected. 
\newedit{Note that, as seen in Eq.~\eqref{eqn:pos_from_vel}, the bias estimate $\mathbf{b}_{a}$ is available in the Girona dataset; however, for our platform, the bias values were found within a small magnitude. Hence, current results are computed without considering bias.}

\subsection{Results and Discussion}

Both datasets were tested using EKF, InEKF, and the proposed C-IPG algorithm. 
\newedit{The covariance matrices for EKF and InEKF were set to $0.1\mathbf{I}$, ensuring uniform variance without cross-correlation.} The parameter values for C-IPG were $N = 5$, $d = 3$, $\alpha = 0.1$, $\delta = 1$, and $K_{0} = 10^{-3} \mathbf{I}$. All algorithms were implemented in \texttt{MATLAB 2024a} and executed on a MacBook Pro with an M1 Pro processor.

\begin{figure}[h!]
    \centering
    \includegraphics[width=0.9\columnwidth, trim = {0 0 0 4cm}, clip]{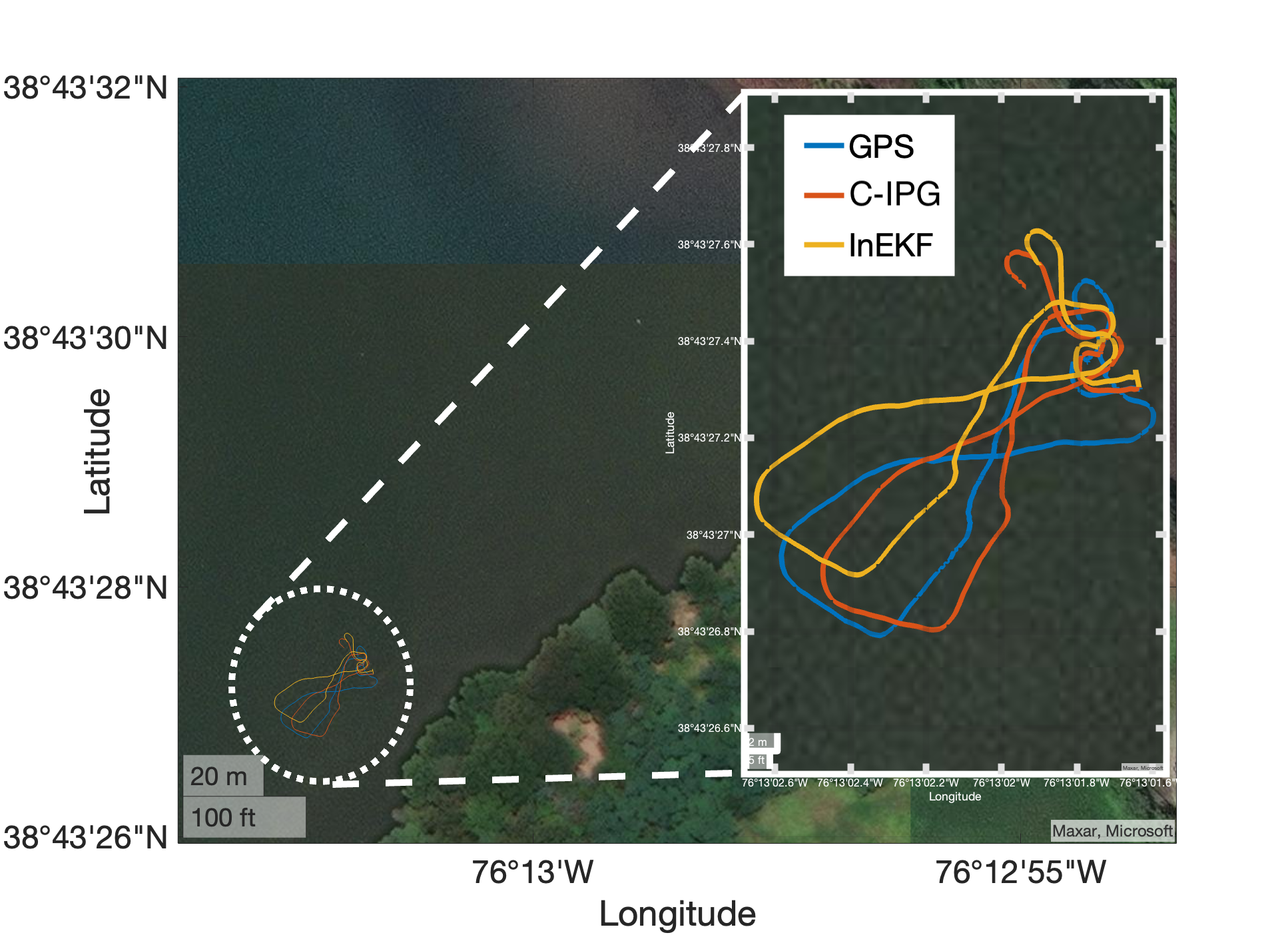}
    \caption{Geographical plot for the estimated long-term trajectory ($> 300$ seconds) observed in the oyster leases (Chesapeake Bay, MD) by the BlueROV2.}
    \label{fig:uuv_geoplot}
\end{figure}

First, the Girona dataset is validated using odometry readings as ground truth. \newedit{The trajectory comparison is shown in Fig.~\ref{fig:girona_traj}, and individual states are compared in Fig.~\ref{fig:uuv_positions_girona}.} Although C-IPG has a higher computational time than EKF or InEKF (Table~\ref{tab:results}), it achieves the best estimation accuracy. Still, the runtime of C-IPG remains below the actual experiment duration, suggesting its feasibility for real-time applications. Similar results were observed in our experiments in the Chesapeake Bay, with individual state comparisons shown in Fig.~\ref{fig:uuv_positions_bluerov}. Additionally, Fig.~\ref{fig:uuv_geoplot} presents the estimated ROV trajectories superimposed on a satellite image, obtained by converting Mercator projection coordinates \cite{snyder1987map} to geographical coordinates for comparison with the trajectory obtained via GPS.

\begin{figure}
    \centering
    \includegraphics[width=\columnwidth]{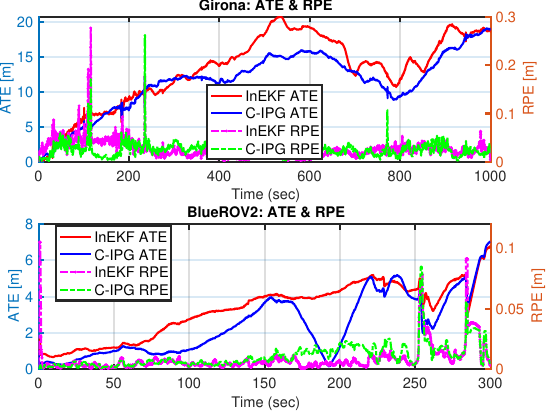}
    \caption{Comparison of Absolute Trajectory Error (ATE) and Relative Pose Error (RPE) over time. While the errors accumulate over time, long-term evaluations show that the observers can maintain relatively accurate estimation results.}
    \label{fig:error_prop}
\end{figure}

Table~\ref{tab:results} summarizes the Mean Absolute Error (MAE) for individual states and the overall Root Mean Squared Error (RMSE) for all three approaches, along with runtime comparisons. RMSE was computed by combining position errors (in meters) with small-angle approximated orientation errors treated as arc lengths in meters, yielding a single metric for pose estimation accuracy. The table includes short-term (100 seconds) and long-term (300 and 1000 seconds) estimates. Short-term results indicate that the platform can handle external sensor outages with total errors of less than one meter. Meanwhile, the long-term estimates suggest that the onboard inertial sensors can achieve relatively accurate estimation for a low-cost system.
\newedit{Hence, the operational time for underwater vehicles to maintain good localization results without surfacing can be selected based on the tolerance level of the estimation error.}
\newedit{Fig.~\ref{fig:error_prop} illustrates the Absolute Trajectory Error (ATE) and Relative Pose Error (RPE)~\cite{8593941} over time, showing that C-IPG maintains lower ATE than InEKF. The RPE remains relatively low for both methods, with occasional spikes likely due to rapid motion or sensor noise.}

\newedit{It is conjectured that higher deviations in the estimate results for Girona dataset are due to the effect of the vehicle's speed. The speed in the Girona dataset is more than twice as fast as that of our ROV. However, a detailed analysis is not yet available and will be investigated in future work. Furthermore, C-IPG outperforms InEKF for a faster convergence after initialization, potentially due to utilizing a moving horizon, which avoids accumulated error in Eq.~\eqref{eqn:pos_from_vel}.}

\section{Conclusion and Future Work}

In this paper, we introduce C-IPG observer, a novel state estimation framework applicable to underwater vehicles and other 6-DOF systems using an inertial sensor suite. Our framework was validated on a public dataset and our test data, demonstrating that it outperforms current filter-based methods in terms of positional accuracy.

\textbf{Limitations: } Although both InEKF and C-IPG show promising results, these stem from observing the initial heading measurement from {AHRS}. Integrating {magnetometers} can contribute to better heading results while initializing and correcting the yaw angle. \newedit{Currently, since only inertial sensors are used, the error increases with time. Other types of sensors (e.g., USBL, pressure, etc.) will be incorporated into the pipeline to further enhance the estimation accuracy.}

\textbf{Future Work:} This work will be extended to prove the observer's convergence and robustness. Further work also includes using this method for observer-based control for a resurfacing or re-localization strategy. A hybrid method combining the optimization accuracy of C-IPG with the computation speed of filter-based methods like InEKF can also be developed for state estimation. Our framework can potentially enhance SLAM algorithm accuracy, especially for those relying on vision or sonar. Addressing accelerometer bias through methods like Allan Variance can improve accuracy.

\section*{ACKNOWLEDGMENTS}
The authors thank Dr. Matthew Gray and Alan Williams at Horn Point Laboratory, UMCES, for their support during field experiments; Dr. Don Webster and Mr. Bobby Leonard for providing access to oyster leases and Mr. Logan Bilbrough for the aerial imagery.

\bibliographystyle{IEEEtran}
\bibliography{refs_ipg_observer, refs_uuv, refs_mhe, refs_rov, refs_imu}

\end{document}